%% file: scis_paper.tex
\documentclass{SCIS2026}
\usepackage{wrapfig}
\begin{document}
\ArticleType{RESEARCH PAPER}
\Year{2025}
\Month{January}
\Vol{68}
\No{1}
\DOI{}
\ArtNo{}
\ReceiveDate{}
\ReviseDate{}
\AcceptDate{}
\OnlineDate{}
\AuthorMark{}
\AuthorCitation{}

\title{CSRA: Controlled Spectral Residual Augmentation for Robust Sepsis Prediction}{Title for citation}


\author[1]{Honglin Guo}{}
\author[1]{Rihao Chang}{}
\author[1]{He Jiao}{}
\author[1]{Weizhi Nie}{weizhinie@tju.edu.cn}
\author[3]{Zhongheng Zhang}{zh\_zhang1984@zju.edu.cn}
\author[2]{Yuehao Shen}{yuehaoshen@163.com}

\address[1]{Tianjin University, Tianjin 300072, China}
\address[2]{Tianjin Medical University General Hospital, Tianjin 300052, China}
\address[3]{Department of Emergency Medicine, Sir Run Run Shaw Hospital, Zhejiang University School of Medicine, Hangzhou 310016, China}
\abstract{
Accurate prediction of future risk and disease progression in sepsis is clinically important for early warning and timely intervention in intensive care. However, short-window sepsis prediction remains challenging, because shorter observation windows provide limited historical evidence, whereas longer prediction horizons reduce the number of patient trajectories with valid future supervision.

To address this problem, we propose CSRA, a Controlled Spectral Residual Augmentation framework for short-window multi-system ICU time series. CSRA first groups variables by clinical systems and extracts system-level and global representations. It then performs input-adaptive residual perturbation in the spectral domain to generate structured and clinically plausible trajectory variations. To improve augmentation stability and controllability, CSRA is trained end-to-end with the downstream predictor under a unified objective, together with anchor consistency loss and controller regularization. Experiments on a MIMIC-IV sepsis cohort across multiple downstream models show that CSRA is consistently competitive and often superior, reducing regression error by 10.2\% in MSE and 3.7\% in MAE over the non-augmentation baseline, while also yielding consistent gains on classification.
CSRA further maintains more favorable performance under shorter observation windows, longer prediction horizons, and smaller training data scales, while also remaining effective on an external clinical dataset~(ZiGongICUinfection), indicating stronger robustness and generalizability in clinically constrained settings.}

\maketitle

\input{sec/1_intro}
\input{sec/2_rel}
\input{sec/3_method}
\input{sec/4_exp}

\input{sec/5_lim}
\input{sec/6_conclusion}

\input{scis_paper.bbl}






\end{document}

%% file: sec/1_intro.tex
\section{Introduction}
\label{introduction}

\begin{wrapfigure}{r}{0.43\columnwidth}
    \vspace{-0.4\baselineskip}
    \centering
    \includegraphics[width=0.40\columnwidth]{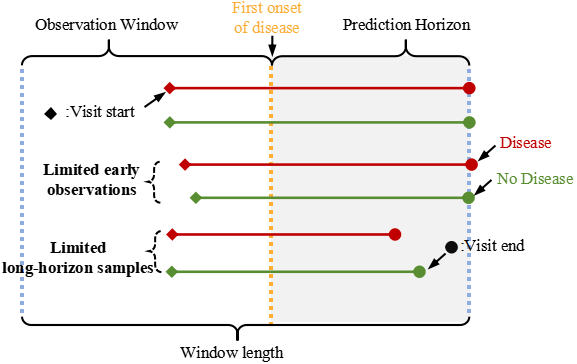}
    \caption{Illustration of temporal robustness challenges. Valid patient trajectories become increasingly limited when observations are farther from sepsis onset or when longer prediction horizons are required.}
    \label{fig:intro_temporal_robustness}
    \vspace{-0.7\baselineskip}
\end{wrapfigure}

Short-window clinical time-series prediction for patients with sepsis is a clinically important yet challenging problem in intensive care~\cite{van2017immunopathology,ferrer2014empiric}. Its clinical value lies in enabling models to infer future risk changes and disease progression from limited recent observations, thereby supporting early warning and timely intervention. Recent advances in deep learning, particularly LSTM- and Transformer-based models~\cite{lstm,hou2020predicting,deshon2022prediction,horng2017creating,kam2017learning,wang2022integrating,zhang2023interpretable,goh2021artificial,liu2019naturalMGP-AttTcn,li2025temporal,yong2024deep,wu2023forecasting}, have substantially improved clinical temporal modeling. However, prediction under shorter observation windows and longer horizons remains intrinsically difficult, because it must be performed with more limited temporal evidence and weaker valid trajectory support~\cite{Multi-TA}.

This degradation reflects limited temporal robustness across different observation windows and prediction horizons, and is mainly driven by two related forms of data scarcity, as illustrated in Fig.~\ref{fig:intro_temporal_robustness}. First, patients often seek medical care only after symptoms become clinically apparent, so observations from earlier stages that are farther from sepsis onset are less frequently available in practice. This restricts the amount of historical evidence accessible to the model. Second, as the prediction horizon increases, the number of patient trajectories covering the corresponding future interval also decreases, since not all visits contain sufficiently long and complete follow-up records. The former limits the temporal evidence contained within each trajectory, whereas the latter reduces training support for diverse disease progression patterns. Together, these factors make short-window and long-horizon prediction substantially more difficult, consistent with prior observations~\cite{Multi-TA,KhoshnevisanC21}.

Data augmentation~\cite{alomar2023data,mohammadi2024deep,iwana2021empirical,yue2022ts2vec,luo2023time,zheng2024parametric,AutoDA-Timeseries} offers a possible way to mitigate this limitation, yet general-purpose time-series augmentation is not well suited to short-window ICU sequences. In this setting, the observed trajectory already contains only limited clinically informative structure, so unconstrained perturbations can easily distort useful signals. In addition, ICU variables are strongly coupled across multiple clinical systems, and arbitrary augmentation may therefore produce physiologically implausible patterns. What is needed is not indiscriminate sample expansion, but controlled and structured augmentation around the original trajectories, so that the model is exposed to more challenging local dynamics without compromising clinical plausibility.

To address this issue, we propose \textbf{CSRA}, a \textbf{C}ontrolled \textbf{S}pectral \textbf{R}esidual \textbf{A}ugmentation framework for short-window, multi-system ICU time series. Our goal is to construct structured augmented samples around the input trajectories while preserving clinically plausible trajectory variations. Specifically, we group input variables by clinical systems and extract both system-level and global representations. Each system trajectory is then mapped into the DCT domain, where residual perturbations are applied over low-, mid-, and high-frequency bands to form controlled deviations from the original trajectories. The augmentation strength is further modulated by the system-level and global representations, allowing perturbations across systems, frequency bands, and temporal locations to vary with the input state rather than being imposed uniformly across samples. Compared with directly modifying the original sequence in the time domain, this design controls augmentation direction and magnitude in a more structured space, making it better suited to short-window clinical time-series modeling.

During training, CSRA is jointly optimized with the downstream predictor under a unified objective. To improve controllability and reduce the risk of unrealistic trajectory generation, we introduce an anchor consistency loss that uses the original branch as a fixed reference for the augmented branch. We further introduce a controller regularization term to prevent inactive or collapsed augmentation patterns and to encourage balanced participation across systems, frequency bands, and internal dimensions. Experimental results across multiple downstream models show that CSRA consistently outperforms representative baselines, reducing regression errors by 10.2\% in MSE and 3.7\% in MAE over the non-augmentation baseline, while also yielding consistent gains on classification. CSRA also remains more effective under shorter observation windows, longer prediction horizons, and smaller training data scales, and generalizes well to an external clinical dataset (\textit{ZiGongICUinfection}~\cite{ZiGongICUinfection}), indicating stronger robustness in information-limited settings.

The main contributions of this work are as follows:
\begin{itemize}
    \item \textbf{Problem perspective:} We investigate short-window sepsis prediction from the perspective of temporal robustness, highlighting the difficulty of maintaining reliable performance when the observation window becomes shorter and the prediction horizon becomes longer.

    \item \textbf{CSRA framework:} We propose CSRA, an end-to-end augmentation framework for short-window multi-system ICU time series, which groups variables by clinical systems and learns input-adaptive spectral residual perturbations to generate structured and clinically plausible trajectory variations.

    \item \textbf{Extensive empirical validation:} Experiments across multiple downstream models show that CSRA consistently improves over the non-augmentation baseline, achieves competitive or superior performance against representative augmentation baselines, remains more effective under shorter windows, longer horizons, and smaller training data scales, and further demonstrates good generalizability on an external clinical dataset (\textit{ZiGongICUinfection}).
\end{itemize}

%% file: sec/2_rel.tex
\section{Related Work}

\subsection{Sepsis Time-Series Modeling}
Given its high mortality in the ICU, sepsis requires accurate early prediction and risk stratification. In recent years, deep learning~\cite{lstm,transformer} has shifted sepsis research from static assessment toward dynamic time-series modeling, substantially advancing early warning and prognostic analysis.

Existing studies~\cite{hou2020predicting,li2025temporal,yong2024deep,wu2023forecasting,raghu2017continuous,choi2024deep} can be broadly grouped into three categories. The first includes traditional statistical and machine learning methods~\cite{hou2020predicting,deshon2022prediction,horng2017creating,kam2017learning,wang2022integrating} based on structured electronic health records. Although relatively interpretable, these approaches remain limited in capturing complex temporal dependencies. The second~\cite{zhang2023interpretable,goh2021artificial,liu2019naturalMGP-AttTcn,li2025temporal,yong2024deep,wu2023forecasting} consists of deep temporal architectures, typically built upon LSTM~\cite{lstm}, Transformer~\cite{transformer}, and their variants, to model long-range dependencies and variable interactions in multivariate clinical sequences. For example, MGP-AttTCN~\cite{liu2019naturalMGP-AttTcn} combines Gaussian-process imputation with attentive temporal convolution to mitigate missingness in clinical data. SCDM~\cite{li2025temporal} improves prediction reliability by causally disentangling sepsis-related factors from confounders, whereas AL-Transformer~\cite{wu2023forecasting} strengthens the modeling of clinical state evolution through an improved sequence modeling mechanism. The third category~\cite{raghu2017continuous,choi2024deep} extends beyond risk prediction to treatment strategy modeling and optimization by jointly considering patient states, treatment actions, and clinical outcomes.

While prior sepsis studies have mainly emphasized model architecture and task formulation, they have paid less attention to the scarcity of clinically critical trajectories under short-window settings. Unlike Multi-TA~\cite{Multi-TA}, which improves robustness through temporal augmentation for early prediction, our work focuses on controllably supplementing key evolving dynamics in short ICU sequences.

\subsection{Time-Series Data Augmentation}

Early studies~\cite{alomar2023data,mohammadi2024deep,iwana2021empirical} on time-series augmentation mainly relied on rule-based transformations in the time domain, such as jittering~\cite{salamon2017deep}, scaling~\cite{pan2020data}, and time warping~\cite{le2016data}. These methods are easy to implement, but they remain largely heuristic and are not well adapted to specific tasks or data distributions.

More recent work~\cite{luo2023time,yue2022ts2vec,zheng2024parametric,AutoDA-Timeseries} has explored two broader directions. One is representation learning~\cite{yue2022ts2vec,luo2023time}, which improves the stability of time-series representations through multiple views or contrastive samples. The other~\cite{zheng2024parametric,chang2024caap,AutoDA-Timeseries} is automated data augmentation (AutoDA), whose central idea is to learn augmentation policies jointly with downstream task optimization. To address the limitations of image-derived AutoDA methods~\cite{cubuk2020randaugment,yang2023survey,A2-aug,yuan2024reaugment,Trivialaugment} in modeling temporal autocorrelation and sequential structure, AutoDA-Timeseries~\cite{AutoDA-Timeseries} proposes a one-stage framework that jointly learns augmentation policies and the downstream model.

Existing augmentation studies~\cite{cubuk2020randaugment,yang2023survey,A2-aug,yuan2024reaugment,Trivialaugment} have mainly focused on general-purpose policy learning, while paying less attention to structural constraints on the augmentation space for short ICU sequences. In contrast, our work emphasizes controlled expansion of key evolving dynamics under short-window clinical settings.

\subsection{Time-Series Analysis Methods}

General time-series modeling has advanced rapidly in recent years. Most existing methods~\cite{zhou2022fedformer,liu2023itransformer,PatchTST,zeng2023transformers,yue2025olinear,wang2024timemixer++,kraus2024xlstm} extend fundamental architectures such as LSTM~\cite{lstm}, linear models, and Transformers~\cite{transformer} to improve the modeling of long-range dependencies, multi-scale patterns, and cross-task generalization. In the medical domain, related studies~\cite{bi2023accurate,ferte2024reservoir,chen2024predictive,jiang2023learning,qin2023t} have further developed task-specific models for dynamic prediction~\cite{bi2023accurate,ferte2024reservoir}, treatment effect assessment~\cite{ye2023web,cao2023estimating}, and early warning~\cite{Multi-TA,ho2023self}. Dedicated frameworks~\cite{li2025mira} tailored to real-world clinical time series have also begun to emerge. Overall, these methods mainly strengthen representational capacity from the model side, while paying less direct attention to performance degradation caused by the scarcity of clinically critical trajectories.

%% file: sec/3_method.tex
\section{Method}

\begin{figure*}[t]
    \centering
    \includegraphics[width=1\textwidth]{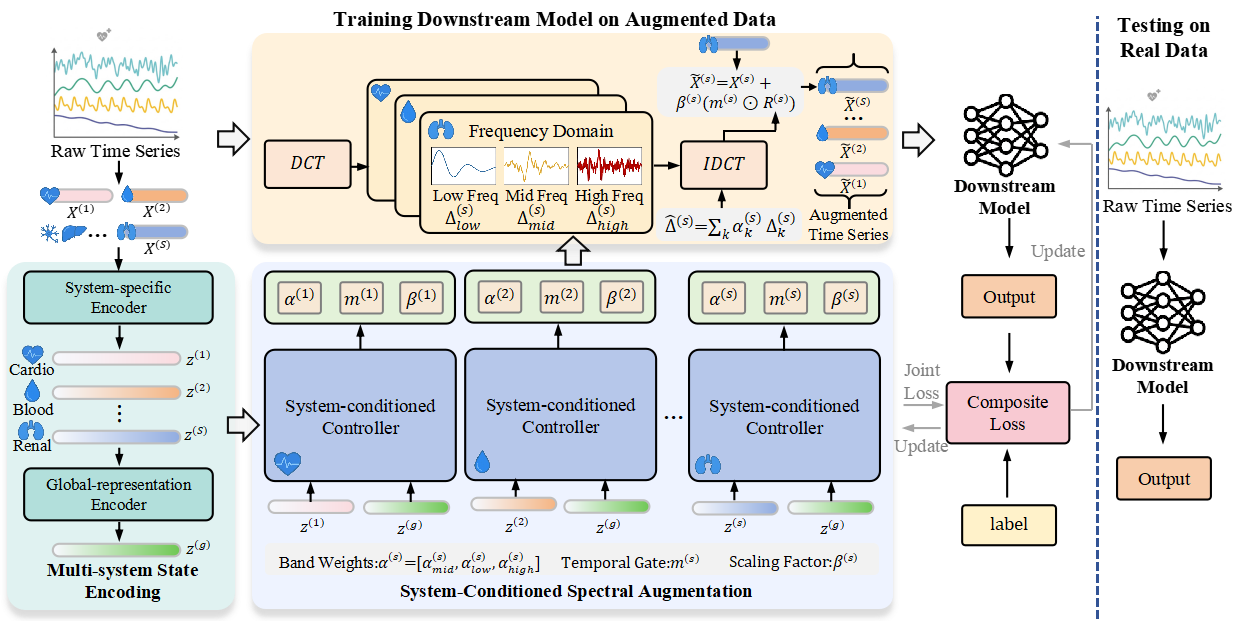}
    \caption{Overall architecture of the \textbf{CSRA} framework. During training, the model first partitions the multivariate time series into multiple clinical systems and extracts the corresponding system-level and global representations. These representations are then used to drive a system-conditioned spectral augmentation module, which generates augmented samples for joint optimization with the downstream model. During inference, only the downstream model is used on the original input, without the augmentation branch.}
    \label{fig.framework}
\end{figure*}

To address data scarcity in short-window sepsis prediction, we propose CSRA, a controlled spectral residual augmentation framework for multi-system ICU time series, as illustrated in Fig.~\ref{fig.framework}. Section~\ref{model:Problem Formulation} defines the prediction tasks and the end-to-end formulation. Section~\ref{model:Multi-system State Encoding} introduces the multi-system state encoding module for extracting system-level and global representations. Section~\ref{model:System-Conditioned Spectral Augmentation} presents the system-conditioned spectral augmentation module, which generates input-adaptive trajectory variations through structured residual perturbation in the frequency domain. Section~\ref{model:Joint Training Objective} describes the unified objective for controllable and stable augmentation.

\subsection{Problem Formulation}
\label{model:Problem Formulation}

In the intensive care setting, each patient sample is represented as a multivariate clinical time series. Let the observed trajectory within a historical window be denoted by:
\begin{equation}
X = [x_1, x_2, \dots, x_W] \in \mathbb{R}^{W \times D},
\end{equation}
where \(W\) is the observation window length, \(D\) is the feature dimension, and \(x_t \in \mathbb{R}^{D}\) denotes the multivariate clinical observation at time step \(t\).

Based on the observed window \(X\), we consider two downstream tasks: regression and classification. For regression, the model predicts future continuous clinical or treatment-related variables at horizon \(H_{\mathrm{reg}}\), written as \(\hat{y}_{\mathrm{reg}} = f^{\mathrm{reg}}_{\theta}(X)\), where \(f^{\mathrm{reg}}_{\theta}(\cdot)\) is the regression predictor parameterized by \(\theta\). For classification, the model predicts future clinical risks or outcomes at horizon \(H_{\mathrm{cls}}\), written as \(\hat{y}_{\mathrm{cls}} = f^{\mathrm{cls}}_{\theta}(X)\), where \(f^{\mathrm{cls}}_{\theta}(\cdot)\) denotes the classification predictor.

To improve downstream learning under limited clinical observations, we introduce an input-adaptive augmentor \(A_{\phi}\), parameterized by \(\phi\), to generate an augmented sample:
\begin{equation}
\tilde{X} = A_{\phi}(X).
\end{equation}
Rather than treating augmentation as a fixed preprocessing step, we jointly optimize the augmentor and the downstream predictor in an end-to-end manner. This allows the augmentation policy to be learned directly under task-level supervision and tailored to the target prediction setting.

\subsection{Multi-system State Encoding}
\label{model:Multi-system State Encoding}

In clinical time series, variables belonging to the same physiological system often exhibit coordinated responses to disease progression and treatment. Motivated by this structure, we partition the input variables into \(S\) predefined clinical systems:
\begin{equation}
X = \{X^{(1)}, X^{(2)}, \dots, X^{(S)}\}, \qquad X^{(s)} \in \mathbb{R}^{W \times D_s},
\end{equation}
where \(D_s\) is the feature dimension of system \(s\) and \(\sum_{s=1}^{S} D_s = D\).

For each system \(s\), a system-specific encoder \(g_s(\cdot)\), followed by temporal pooling, produces a local representation \(z^{(s)} = \mathrm{Pool}(g_s(X^{(s)}))\). The local representations are then aggregated to form a global representation:
\begin{equation}
z^{g} = h\!\left([z^{(1)}; z^{(2)}; \cdots; z^{(S)}]\right),
\end{equation}
where \(h(\cdot)\) denotes a linear projection and \([\cdot;\cdot]\) denotes concatenation. We denote the resulting hierarchical state set as \(Z = \{z^{(1)}, z^{(2)}, \dots, z^{(S)}, z^{g}\}\).

This hierarchical representation provides the conditioning context for augmentation. The system-level representations preserve local physiological structure, whereas the global representation summarizes the overall patient state and supports coordinated modulation across systems.

\subsection{System-Conditioned Spectral Augmentation}
\label{model:System-Conditioned Spectral Augmentation}

Under short-window settings, direct perturbation in the time domain can easily distort the limited informative structure contained in the observed trajectory. We therefore perform augmentation in the frequency domain, where sequence variation can be modulated in a more structured manner.

For each clinical system \(s\), we apply the discrete cosine transform (DCT) along the temporal dimension:
\begin{equation}
\hat{X}^{(s)} = F\!\left(X^{(s)}\right),
\end{equation}
where \(F(\cdot)\) denotes the DCT. We then decompose \(\hat{X}^{(s)}\) into three band-specific components:
\begin{equation}
\hat{X}^{(s)} = \Delta^{(s)}_{\mathrm{low}} + \Delta^{(s)}_{\mathrm{mid}} + \Delta^{(s)}_{\mathrm{high}},
\end{equation}
where \(\Delta^{(s)}_{k}\) denotes the spectral component of system \(s\) within band \(k \in \{\mathrm{low}, \mathrm{mid}, \mathrm{high}\}\). The low-frequency component mainly captures slow-varying trends, the mid-frequency component captures intermediate-scale fluctuations, and the high-frequency component captures rapid local variations.

To adapt the perturbation strength to patient state, system identity, frequency band, and temporal location, we concatenate the local and global representations and feed them into a controller:
\begin{equation}
\alpha^{(s)},\, m^{(s)},\, \beta^{(s)} = \Psi\!\left([z^{(s)}; z^{g}]\right),
\end{equation}
where \(\Psi(\cdot)\) denotes the controller, \(\alpha^{(s)} = [\alpha^{(s)}_{\mathrm{low}}, \alpha^{(s)}_{\mathrm{mid}}, \alpha^{(s)}_{\mathrm{high}}]\) controls the relative contribution of the three frequency bands, \(m^{(s)} \in [0,1]^W\) is a temporal gate, and \(\beta^{(s)} \in (0,1)\) is a system-level scaling factor.

The three spectral components are first reweighted as:
\begin{equation}
\bar{\Delta}^{(s)} =
\alpha^{(s)}_{\mathrm{low}} \Delta^{(s)}_{\mathrm{low}} +
\alpha^{(s)}_{\mathrm{mid}} \Delta^{(s)}_{\mathrm{mid}} +
\alpha^{(s)}_{\mathrm{high}} \Delta^{(s)}_{\mathrm{high}}.
\end{equation}
The weighted perturbation is then mapped back to the time domain as \(R^{(s)} = F^{-1}(\bar{\Delta}^{(s)})\), where \(F^{-1}(\cdot)\) denotes the inverse DCT. We further apply the temporal gate and the system-level scaling factor to obtain the augmented system trajectory:
\begin{equation}
\tilde{X}^{(s)} = X^{(s)} + \beta^{(s)} \left(m^{(s)} \odot R^{(s)}\right),
\end{equation}
where \(\odot\) denotes element-wise multiplication with temporal broadcasting over the feature dimension. The final augmented sample is obtained by merging all augmented systems, i.e., \(\tilde{X} = \mathrm{Merge}(\tilde{X}^{(1)}, \tilde{X}^{(2)}, \dots, \tilde{X}^{(S)})\).

This design decouples frequency-band reweighting from time-domain scaling, allowing augmentation to remain structured, input-adaptive, and system-aware.

\subsection{Joint Training Objective}
\label{model:Joint Training Objective}

To preserve task supervision while keeping augmentation controllable, we optimize the joint objective:
\begin{equation}
L_{\mathrm{total}} = L_{\mathrm{orig}} + \lambda_{\mathrm{cons}} L_{\mathrm{cons}} + \lambda_{\mathrm{ctrl}} L_{\mathrm{ctrl}},
\end{equation}
where \(L_{\mathrm{orig}}\) is the original task loss, \(L_{\mathrm{cons}}\) is the anchor consistency loss, and \(L_{\mathrm{ctrl}}\) is the controller regularization loss. The coefficients \(\lambda_{\mathrm{cons}}\) and \(\lambda_{\mathrm{ctrl}}\) balance task supervision, augmentation stability, and augmentation control. For regression, \(L_{\mathrm{orig}}\) is instantiated as mean squared error; for classification, it is instantiated as cross-entropy loss.

To stabilize augmentation and reduce the risk of unrealistic trajectory variation, we use the prediction of the original branch as a fixed anchor and impose a consistency constraint on the augmented branch:
\begin{equation}
L_{\mathrm{cons}} =
\mathrm{SmoothL1}\!\left(f_{\theta}(\tilde{X}),\, \mathrm{sg}(f_{\theta}(X))\right),
\end{equation}
where \(f_{\theta}(\cdot)\) denotes the downstream predictor and \(\mathrm{sg}(\cdot)\) denotes the stop-gradient operation.

To prevent the controller from producing overly strong, overly concentrated, or inactive modulation patterns, we further regularize it as:
\begin{equation}
L_{\mathrm{ctrl}} = \lambda_{\mathrm{inter}} L_{\mathrm{inter}} + \lambda_{\mathrm{intra}} L_{\mathrm{intra}},
\end{equation}
where \(L_{\mathrm{inter}}\) constrains the overall modulation strength across systems and frequency bands, and \(L_{\mathrm{intra}}\) regularizes the temporal participation pattern induced by the gate.

For system \(s\) and frequency band \(k \in \{\mathrm{low}, \mathrm{mid}, \mathrm{high}\}\), we define the effective modulation strength as:
\begin{equation}
u^{(s)}_{k} = \beta^{(s)} \alpha^{(s)}_{k} \left\| \Delta^{(s)}_{k} \right\|_{1}.
\end{equation}
The inter-controller regularization is then given by:
\begin{equation}
L_{\mathrm{inter}} =
\sum_{s=1}^{S} \sum_{k}
\mathrm{softplus}\!\left(u^{(s)}_{k} - \tau\right)^2,
\end{equation}
where \(\tau\) denotes the upper bound of the desired modulation strength. This term suppresses excessively strong local modulation and prevents a small subset of systems or bands from dominating the augmentation process.

We further regularize the temporal participation pattern of the gate. For system \(s\), the normalized temporal participation weight at time step \(t\) is defined as:
\begin{equation}
q^{(s)}_{t} =
\frac{m^{(s)}_{t}}{\sum_{j=1}^{W} m^{(s)}_{j} + \varepsilon},
\end{equation}
where \(\varepsilon\) is a numerical stability constant. Let \(q^{(s)} = [q^{(s)}_{1}, \dots, q^{(s)}_{W}]\) denote the resulting temporal participation distribution. The corresponding intra-controller regularization is:
\begin{equation}
L_{\mathrm{intra}} =
-\frac{1}{S}\sum_{s=1}^{S} H\!\left(q^{(s)}\right),
\end{equation}
where \(H(\cdot)\) denotes entropy. This term discourages the temporal gate from concentrating most perturbation mass on only a few time steps, thereby encouraging broader temporal participation.

The final objective unifies task supervision, augmentation stability, and controller regularization within a single end-to-end training framework.

%% file: sec/4_exp.tex
\section{Experiments} 

\subsection{Implementation}

All experiments were run on four NVIDIA GeForce RTX 4090 GPUs (24GB each). The cohort was randomly divided into training, validation, and test sets with a ratio of 7:1.5:1.5. Unless otherwise stated, the default setting used an observation window of \(W=6\), a regression horizon of \(H_{\mathrm{reg}}=1\), and a classification horizon of \(H_{\mathrm{cls}}=6\).

Model optimization was performed using Adam, with a learning rate of \(1\times10^{-3}\), weight decay of \(1\times10^{-4}\), dropout of \(0.1\), and batch size of \(64\). Training continued for at most \(100\) epochs, and early stopping with a patience of \(10\) was applied according to the validation loss.

For hierarchical state encoding, the clinical variables were grouped into 9 predefined physiological systems based on clinical priors, including the respiratory and circulatory systems. In the joint objective, we set \(\lambda_{\mathrm{cons}}=0.5\) and \(\lambda_{\mathrm{ctrl}}=0.1\). Within the controller regularization term, \(\lambda_{\mathrm{inter}}=1.0\), \(\lambda_{\mathrm{intra}}=0.1\), \(\tau=0.5\), and \(\varepsilon=10^{-6}\).

\subsection{Evaluation Benchmarks}

\subsubsection{Datasets}

Our experiments follow the MIMIC-Sepsis~\cite{MIMIC-Sepsis} processing pipeline, from which we derived a sepsis cohort of 34,793 patients from MIMIC-IV. Each sample is represented as a clinical trajectory with 63 variables. The timeline is not anchored at hospital admission; instead, it is aligned to a sepsis-related time point and resampled at a fixed 4-hour resolution.

\subsubsection{Benchmark Tasks}

We consider two prediction tasks: regression and classification. Given an observation window \(W\), the regression task predicts future continuous clinical or treatment-related variables at horizon \(H_{\mathrm{reg}}\), whereas the classification task predicts the occurrence of target risk events or clinical outcomes within horizon \(H_{\mathrm{cls}}\). For classification, we consider three outcomes: 90-day mortality, readmission, and septic shock.

\subsubsection{Evaluation Metrics}

For regression, we use MAE and MSE to quantify prediction error. For classification, given the substantial class imbalance in clinical outcomes, we report AUROC and AUPRC as the primary metrics. Unless otherwise stated, all results are reported as mean \(\pm\) standard deviation over five random seeds.

\begin{table*}[tp]
\centering
\caption{Comparison of regression performance across baselines and Ours. Best results are in \textbf{bold}, and the second-best are underlined.}
\label{table:regression_results}
\resizebox{0.98\linewidth}{!}{%
\begin{tabular}{l|c|ccccccc}
\toprule
\multirow{2}{*}{\textbf{Downstream Model}} & \multirow{2}{*}{\textbf{Metrics}} & \multicolumn{7}{c}{\textbf{Methods}} \\
\cmidrule(lr){3-9}
 & & \textbf{NoAug} & \textbf{InfoTS} & \textbf{AutoTCL} & \textbf{TrivialAugment} & \textbf{A2Aug} & \textbf{AutoDA-Timeseries} & \textbf{Ours} \\
\midrule
\multirow{2}{*}{Linear}
& MSE & 0.172($\pm$.002) & 0.178($\pm$.005) & 0.176($\pm$.004) & 0.168($\pm$.003) & \underline{0.160($\pm$.003)} & 0.163($\pm$.004) & \textbf{0.157($\pm$.002)} \\
& MAE & 0.177($\pm$.002) & 0.181($\pm$.004) & 0.179($\pm$.003) & 0.176($\pm$.002) & \underline{0.174($\pm$.002)} & 0.175($\pm$.003) & \textbf{0.173($\pm$.001)} \\
\midrule
\multirow{2}{*}{LSTM}
& MSE & 0.164($\pm$.007) & 0.170($\pm$.006) & 0.167($\pm$.005) & 0.156($\pm$.004) & 0.149($\pm$.004) & \textbf{0.143($\pm$.003)} & \underline{0.144($\pm$.002)} \\
& MAE & 0.174($\pm$.001) & 0.178($\pm$.003) & 0.176($\pm$.002) & 0.171($\pm$.002) & 0.169($\pm$.003) & \underline{0.166($\pm$.002)} & \textbf{0.165($\pm$.001)} \\
\midrule
\multirow{2}{*}{Transformer}
& MSE & 0.202($\pm$.010) & 0.208($\pm$.010) & 0.204($\pm$.009) & 0.196($\pm$.008) & 0.189($\pm$.007) & \underline{0.184($\pm$.007)} & \textbf{0.182($\pm$.006)} \\
& MAE & 0.222($\pm$.005) & 0.226($\pm$.006) & 0.224($\pm$.005) & 0.221($\pm$.005) & 0.219($\pm$.004) & \underline{0.216($\pm$.004)} & \textbf{0.214($\pm$.003)} \\
\bottomrule
\end{tabular}%
}
\end{table*}
\begin{table*}[tp]
\centering
\caption{Comparison of classification performance across baselines and Ours. Best results are in \textbf{bold}, and the second-best are underlined.}
\label{table:classification_results}
\resizebox{0.98\linewidth}{!}{%
\begin{tabular}{l|c|ccccccc}
\toprule
\multirow{2}{*}{\textbf{Downstream Model}} & \multirow{2}{*}{\textbf{Metrics}} & \multicolumn{7}{c}{\textbf{Methods}} \\
\cmidrule(lr){3-9}
 & & \textbf{NoAug} & \textbf{InfoTS} & \textbf{AutoTCL} & \textbf{TrivialAugment} & \textbf{A2Aug} & \textbf{AutoDA-Timeseries} & \textbf{Ours} \\
\midrule
\multirow{2}{*}{Linear}
& AUROC & 0.884($\pm$.010) & 0.889($\pm$.011) & 0.891($\pm$.009) & 0.888($\pm$.010) & \underline{0.894($\pm$.008)} & 0.893($\pm$.007) & \textbf{0.895($\pm$.006)} \\
& AUPRC & 0.701($\pm$.011) & 0.708($\pm$.012) & 0.711($\pm$.010) & 0.707($\pm$.011) & \underline{0.716($\pm$.009)} & 0.714($\pm$.007) & \textbf{0.717($\pm$.008)} \\
\midrule
\multirow{2}{*}{LSTM}
& AUROC & 0.889($\pm$.009) & 0.894($\pm$.010) & 0.897($\pm$.008) & 0.893($\pm$.009) & \textbf{0.902($\pm$.007)} & \underline{0.900($\pm$.006)} & 0.899($\pm$.006) \\
& AUPRC & 0.713($\pm$.013) & 0.720($\pm$.012) & 0.724($\pm$.011) & 0.719($\pm$.011) & \underline{0.731($\pm$.010)} & 0.730($\pm$.009) & \textbf{0.733($\pm$.008)} \\
\midrule
\multirow{2}{*}{Transformer}
& AUROC & 0.893($\pm$.008) & 0.898($\pm$.008) & 0.899($\pm$.007) & 0.897($\pm$.008) & 0.904($\pm$.006) & \underline{0.905($\pm$.005)} & \textbf{0.906($\pm$.004)} \\
& AUPRC & 0.721($\pm$.010) & 0.728($\pm$.009) & 0.732($\pm$.009) & 0.727($\pm$.010) & \underline{0.739($\pm$.008)} & 0.738($\pm$.007) & \textbf{0.741($\pm$.006)} \\
\bottomrule
\end{tabular}%
}
\end{table*}

\subsection{Comparison Experiments}

\subsubsection{Comparison Methods}

We compare CSRA with two groups of baselines. The first is the non-augmentation setting, denoted as NoAug, in which the downstream model is trained directly on the original data. The second includes two-stage representation learning methods, namely InfoTS~\cite{luo2023time} and AutoTCL~\cite{zheng2024parametric}, both of which construct contrastive views to learn task-agnostic temporal representations. We further include three representative automated augmentation methods, namely TrivialAugment~\cite{Trivialaugment}, A2Aug~\cite{A2-aug}, and AutoDA-Timeseries~\cite{AutoDA-Timeseries}.

\subsubsection{Downstream Models}

To evaluate whether the effect of CSRA is consistent across different predictive architectures, we adopt three representative backbone models widely used in both general and clinical time-series prediction:
1) \textbf{Linear}~\cite{linear}, which flattens the input sequence within the observation window and generates predictions through a linear layer, with hidden dimension 128;
2) \textbf{LSTM}~\cite{lstm}, which uses a two-layer stacked recurrent architecture to capture temporal dependencies, with hidden dimension 128;
3) \textbf{Transformer}~\cite{transformer}, which uses a single-layer encoder with hidden dimension 128 and 4 attention heads.

\subsubsection{Quantitative Comparisons}

Tables~\ref{table:regression_results} and \ref{table:classification_results} summarize the quantitative results on the regression and classification tasks, respectively. Across different downstream models, CSRA achieves the best or near-best performance in most settings, indicating that the proposed augmentation framework consistently improves short-window sepsis prediction for both continuous-state forecasting and risk prediction.

The gains are more evident on the regression task. Relative to the non-augmentation baseline, the average MSE across the three downstream models decreases from 0.179 to 0.161, and the average MAE decreases from 0.191 to 0.184, corresponding to relative reductions of 10.2\% and 3.7\%, respectively. CSRA attains the best result in 5 of the 6 regression settings. This pattern suggests that CSRA provides more effective training support for modeling continuous state evolution under short-window conditions. On both the Linear~\cite{linear} and Transformer~\cite{transformer} backbones, CSRA outperforms all competing methods on both MSE and MAE, indicating that the proposed multi-system spectral residual augmentation is particularly suitable for continuous-value prediction.

CSRA also improves classification performance in a consistent manner. Compared with the non-augmentation baseline, the average AUROC increases from 0.889 to 0.900, while the average AUPRC rises from 0.712 to 0.730. Although the margins are smaller than those observed on regression, CSRA still achieves the best overall classification performance in most settings. This result suggests that the proposed augmentation remains beneficial for short-window risk prediction.

A comparison across method categories reveals a clear pattern. The two-stage representation learning methods, including InfoTS~\cite{luo2023time} and AutoTCL~\cite{zheng2024parametric}, yield only limited gains overall. TrivialAugment~\cite{Trivialaugment}, A2Aug~\cite{A2-aug}, and AutoDA-Timeseries~\cite{AutoDA-Timeseries} provide more consistent improvements over NoAug, but they are still surpassed by CSRA in most settings. These findings suggest that, for short-window ICU time series, neither generic augmentation policies nor two-stage representation transfer is sufficient to adequately enrich clinically relevant temporal variations. By contrast, CSRA combines multi-system modeling, spectral residual perturbation, and jointly optimized regularization to expand the training distribution in a more structured and clinically compatible manner, leading to more stable gains on both regression and classification tasks.

\begin{table*}[tp]
\centering
\caption{Regression results of integrating CSRA into three representative downstream models, including OLinear~\cite{yue2025olinear}, TimeMixer++~\cite{wang2024timemixer++}, and AL-Transformer~\cite{wu2023forecasting}. Best results are in \textbf{bold}, and the second-best are underlined.}
\label{table:compatibility_regression}
\resizebox{0.98\linewidth}{!}{%
\begin{tabular}{l|c|ccccccc}
\toprule
\multirow{2}{*}{\textbf{Downstream Model}} & \multirow{2}{*}{\textbf{Metrics}} & \multicolumn{7}{c}{\textbf{Methods}} \\
\cmidrule(lr){3-9}
 & & \textbf{NoAug} & \textbf{InfoTS} & \textbf{AutoTCL} & \textbf{TrivialAugment} & \textbf{A2Aug} & \textbf{AutoDA-Timeseries} & \textbf{Ours} \\
\midrule
\multirow{2}{*}{OLinear}
& MSE & 0.173($\pm$.007) & 0.171($\pm$.009) & 0.169($\pm$.008) & 0.166($\pm$.007) & \underline{0.162($\pm$.008)} & 0.163($\pm$.007) & \textbf{0.159($\pm$.010)} \\
& MAE & 0.177($\pm$.004) & 0.179($\pm$.005) & 0.178($\pm$.005) & 0.177($\pm$.004) & 0.176($\pm$.004) & \underline{0.175($\pm$.004)} & \textbf{0.174($\pm$.003)} \\
\midrule
\multirow{2}{*}{TimeMixer++}
& MSE & 0.168($\pm$.002) & 0.165($\pm$.004) & 0.163($\pm$.003) & 0.159($\pm$.004) & 0.154($\pm$.005) & \underline{0.152($\pm$.004)} & \textbf{0.150($\pm$.006)} \\
& MAE & 0.189($\pm$.001) & 0.188($\pm$.003) & 0.188($\pm$.002) & 0.187($\pm$.003) & 0.187($\pm$.003) & \underline{0.186($\pm$.002)} & \textbf{0.185($\pm$.003)} \\
\midrule
\multirow{2}{*}{AL-Transformer}
& MSE & 0.176($\pm$.009) & 0.171($\pm$.008) & 0.168($\pm$.008) & 0.160($\pm$.007) & 0.152($\pm$.006) & \underline{0.148($\pm$.006)} & \textbf{0.144($\pm$.005)} \\
& MAE & 0.169($\pm$.002) & 0.167($\pm$.003) & 0.166($\pm$.002) & 0.165($\pm$.002) & 0.163($\pm$.002) & \underline{0.162($\pm$.001)} & \textbf{0.161($\pm$.001)} \\
\bottomrule
\end{tabular}%
}
\end{table*}

\begin{table*}[tp]
\centering
\caption{Classification results of integrating CSRA into three representative downstream  models, including OLinear~\cite{yue2025olinear}, TimeMixer++~\cite{wang2024timemixer++}, and AL-Transformer~\cite{wu2023forecasting}. Best results are in \textbf{bold}, and the second-best are underlined.}
\label{table:compatibility_classification}
\resizebox{0.98\linewidth}{!}{%
\begin{tabular}{l|c|ccccccc}
\toprule
\multirow{2}{*}{\textbf{Downstream Model}} & \multirow{2}{*}{\textbf{Metrics}} & \multicolumn{7}{c}{\textbf{Methods}} \\
\cmidrule(lr){3-9}
 & & \textbf{NoAug} & \textbf{InfoTS} & \textbf{AutoTCL} & \textbf{TrivialAugment} & \textbf{A2Aug} & \textbf{AutoDA-Timeseries} & \textbf{Ours} \\
\midrule
\multirow{2}{*}{OLinear}
& AUROC & 0.901($\pm$.009) & 0.903($\pm$.009) & 0.904($\pm$.008) & 0.904($\pm$.007) & \textbf{0.907($\pm$.007)} & \underline{0.906($\pm$.006)} & \textbf{0.907($\pm$.006)} \\ 
& AUPRC & 0.737($\pm$.011) & 0.741($\pm$.012) & 0.743($\pm$.010) & 0.744($\pm$.010) & \underline{0.747($\pm$.009)} & 0.746($\pm$.008) & \textbf{0.748($\pm$.007)} \\
\midrule
\multirow{2}{*}{TimeMixer++}
& AUROC & 0.902($\pm$.007) & 0.904($\pm$.008) & 0.905($\pm$.007) & 0.905($\pm$.006) & 0.906($\pm$.006) & \underline{0.907($\pm$.005)} & \textbf{0.908($\pm$.005)} \\
& AUPRC & 0.740($\pm$.012) & 0.743($\pm$.011) & 0.745($\pm$.011) & 0.746($\pm$.011) & \underline{0.750($\pm$.009)} & \textbf{0.751($\pm$.009)} & \underline{0.750($\pm$.008)} \\ 
\midrule
\multirow{2}{*}{AL-Transformer}
& AUROC & 0.900($\pm$.008) & 0.903($\pm$.008) & 0.904($\pm$.007) & 0.905($\pm$.006) & 0.907($\pm$.006) & \underline{0.908($\pm$.005)} & \textbf{0.909($\pm$.004)} \\
& AUPRC & 0.739($\pm$.012) & 0.744($\pm$.011) & 0.746($\pm$.010) & 0.748($\pm$.011) & \underline{0.752($\pm$.009)} & 0.751($\pm$.008) & \textbf{0.754($\pm$.007)} \\ 
\bottomrule
\end{tabular}%
}
\vspace{-5pt}
\end{table*}

\subsection{Model Compatibility}

To further examine the generality of CSRA, we integrate it with three stronger downstream  models, including two recent general-purpose time-series models, OLinear~\cite{yue2025olinear} and TimeMixer++~\cite{wang2024timemixer++}, as well as AL-Transformer~\cite{wu2023forecasting}, which is designed specifically for sepsis prediction. Tables~\ref{table:compatibility_regression} and \ref{table:compatibility_classification} report the corresponding results on the regression and classification tasks.

Across all three downstream  models, CSRA remains consistently competitive. On the regression task, it achieves the best performance for every metric--model combination. For instance, when integrated with AL-Transformer, CSRA reduces the MSE from 0.176 to 0.144 relative to the non-augmentation baseline, while also attaining the lowest MAE. On the classification task, CSRA likewise achieves the best or near-best performance across different downstream  models, with the most visible gains observed on AL-Transformer and TimeMixer++. These results indicate that the efficacy of CSRA is not confined to a specific predictor architecture, but transfers well across both general-purpose time-series models and a sepsis-oriented backbone.

\subsection{Ablation Study}

To assess the contribution of the main components in CSRA, we conduct module-level ablation experiments by removing each component individually while keeping the remaining settings unchanged. For fairness, each ablated variant is implemented as a simplified counterpart that remains compatible with the overall framework.

Specifically, in \textbf{w/o Multi-system State Encoding}, we remove the clinical-system partition and the associated system-level conditioning, and instead apply unified augmentation over the full variable set to examine the contribution of the system-aware design. In \textbf{w/o Spectral Augmentation}, we replace spectral residual augmentation with direct residual perturbation in the time domain to evaluate the role of frequency-domain augmentation. In \textbf{w/o Composite Loss}, we retain only the original task loss \(\mathcal{L}_{\mathrm{orig}}\) and remove both the anchor consistency loss and the controller regularization term, in order to assess the role of the joint objective in controllable augmentation.

Table~\ref{table:ablation_results} shows that all three components contribute to the final performance. Overall, the complete CSRA achieves the best or near-best results in most settings, whereas each ablated variant leads to performance drops on at least part of the regression and classification metrics. Among them, removing the composite loss causes the most consistent degradation across downstream  models, highlighting the role of the joint objective in stabilizing the augmentation process. The performance declines caused by w/o Multi-system State Encoding and w/o Spectral Augmentation are also apparent in most settings, especially on regression metrics, further supporting the value of system-aware modeling and frequency-domain residual perturbation in the full framework.

\begin{table*}[tp]
\centering
\caption{Ablation results on both regression and classification tasks. ``w/o Multi-system'' removes the multi-system state encoding and applies unified augmentation over all variables; ``w/o Spectral-Aug'' replaces spectral residual augmentation with direct time-domain residual perturbation; ``w/o Composite Loss'' removes the anchor consistency and controller regularization terms, retaining only the original task loss. Best results are in \textbf{bold}, and the second-best are underlined.}
\label{table:ablation_results}
\resizebox{0.8\linewidth}{!}{%
\begin{tabular}{l|c|c|ccccc}
\toprule
\multirow{2}{*}{\textbf{Downstream Model}} & \multirow{2}{*}{\textbf{Task}} & \multirow{2}{*}{\textbf{Metric}} & \multicolumn{5}{c}{\textbf{Methods}} \\
\cmidrule(lr){4-8}
& & & \textbf{NoAug} & \textbf{w/o Multi-system} & \textbf{w/o Spectral-Aug} & \textbf{w/o Composite Loss} & \textbf{Ours} \\
\midrule
\multirow{4}{*}{Linear}
& \multirow{2}{*}{Reg.}
& MSE   & 0.172($\pm$.002) & \underline{0.161($\pm$.003)} & 0.163($\pm$.004) & 0.166($\pm$.004) & \textbf{0.157($\pm$.002)} \\
& & MAE   & 0.177($\pm$.002) & \textbf{0.172($\pm$.002)} & 0.174($\pm$.002) & 0.177($\pm$.003) & \underline{0.173($\pm$.001)} \\
\cmidrule(lr){2-8}
& \multirow{2}{*}{Cls.}
& AUROC & 0.884($\pm$.010) & 0.891($\pm$.008) & \underline{0.892($\pm$.008)} & 0.888($\pm$.009) & \textbf{0.895($\pm$.006)} \\
& & AUPRC & 0.701($\pm$.021) & \underline{0.713($\pm$.014)} & 0.712($\pm$.015) & 0.708($\pm$.016) & \textbf{0.717($\pm$.008)} \\
\midrule
\multirow{4}{*}{LSTM}
& \multirow{2}{*}{Reg.}
& MSE   & 0.164($\pm$.007) & \underline{0.147($\pm$.003)} & 0.149($\pm$.004) & 0.153($\pm$.004) & \textbf{0.144($\pm$.002)} \\
& & MAE   & 0.174($\pm$.001) & \underline{0.167($\pm$.002)} & 0.168($\pm$.002) & 0.169($\pm$.002) & \textbf{0.165($\pm$.001)} \\
\cmidrule(lr){2-8}
& \multirow{2}{*}{Cls.}
& AUROC & 0.889($\pm$.009) & \underline{0.897($\pm$.007)} & 0.898($\pm$.007) & 0.893($\pm$.008) & \textbf{0.899($\pm$.006)} \\
& & AUPRC & 0.713($\pm$.020) & 0.727($\pm$.014) & \underline{0.729($\pm$.014)} & 0.723($\pm$.015) & \textbf{0.733($\pm$.008)} \\
\midrule
\multirow{4}{*}{Transformer}
& \multirow{2}{*}{Reg.}
& MSE   & 0.202($\pm$.010) & 0.186($\pm$.007) & \underline{0.185($\pm$.007)} & 0.190($\pm$.008) & \textbf{0.182($\pm$.006)} \\
& & MAE   & 0.222($\pm$.005) & 0.216($\pm$.004) & \underline{0.215($\pm$.004)} & 0.218($\pm$.004) & \textbf{0.214($\pm$.003)} \\
\cmidrule(lr){2-8}
& \multirow{2}{*}{Cls.}
& AUROC & 0.893($\pm$.008) & \underline{0.903($\pm$.006)} & 0.902($\pm$.006) & 0.899($\pm$.007) & \textbf{0.906($\pm$.004)} \\
& & AUPRC & 0.721($\pm$.019) & \textbf{0.743($\pm$.008)} & \underline{0.741($\pm$.012)} & 0.733($\pm$.014) & \underline{0.741($\pm$.006)} \\
\bottomrule
\end{tabular}%
}
\end{table*}

\subsection{Effect of Training Data Scale}

\begin{figure*}[t!]
    \centering
    \includegraphics[width=0.98\textwidth]{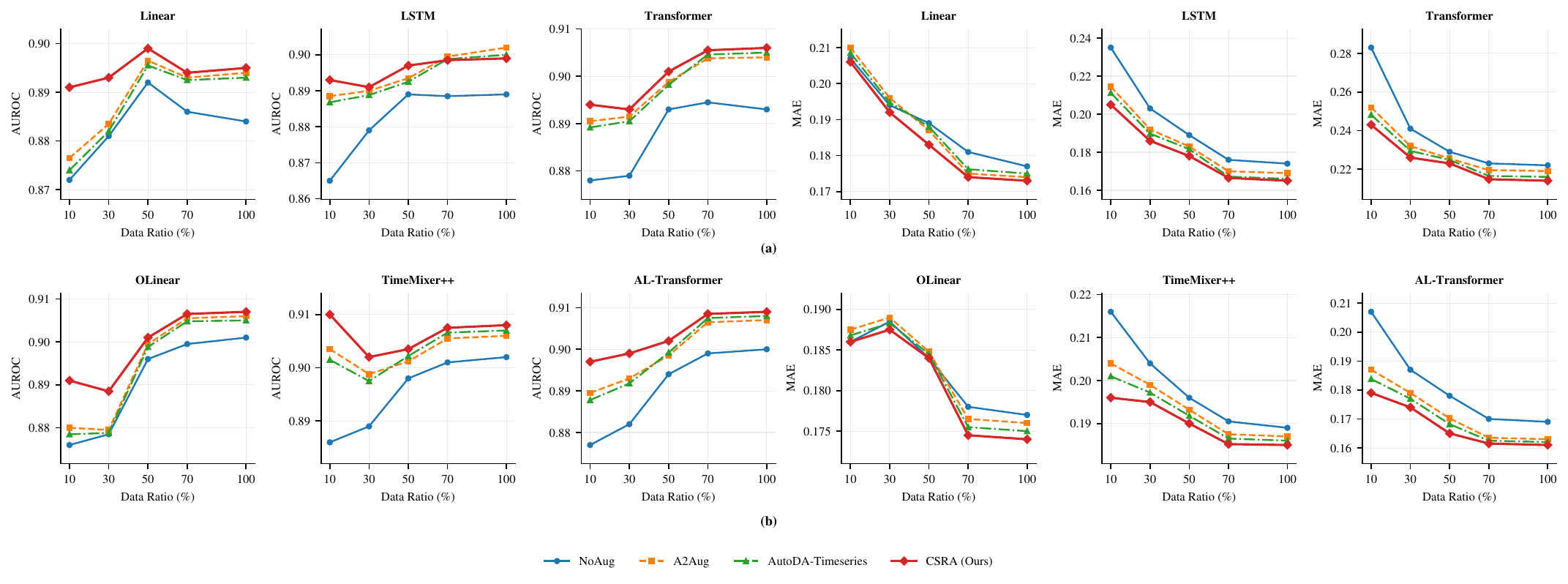}
    \caption{Performance comparison under different training data ratios on six downstream models, where (a) shows the results for the models used in the comparison experiments and (b) shows the results for those used in the model compatibility experiments. The results show that CSRA maintains more favorable performance, especially under smaller training data scales.}
    \label{fig:data_ratio_analysis}
\end{figure*}

To examine the behavior of CSRA under limited training data, we vary the training-set ratio from 10\% to 100\% and compare CSRA with the two strongest baselines, A2Aug~\cite{A2-aug} and AutoDA-Timeseries~\cite{AutoDA-Timeseries}. Figure~\ref{fig:data_ratio_analysis} reports the results on six downstream models, using AUROC for classification and MAE for regression.

As the amount of training data decreases, the performance of all methods deteriorates to varying degrees. Even so, CSRA remains the best or among the best methods in most settings, and its advantage becomes more visible at the lower data ratios of 10\% and 30\%. This pattern suggests that the proposed augmentation strategy provides more effective support when the number of available training samples is limited.

The two tasks also show different sensitivities to data scale. For classification, AUROC generally improves as the training ratio increases, and CSRA remains consistently competitive across most downstream  models. For regression, MAE is more sensitive to data reduction, with the error increasing more noticeably at lower data ratios. Despite this, CSRA still achieves the lowest error in most cases. Taken together, these results indicate that CSRA maintains stronger performance across different data scales, with the clearest gains appearing in the more data-constrained settings.

\subsection{Temporal Robustness Analysis}

\begin{figure*}[!th]
    \centering
    \includegraphics[width=0.98\textwidth]{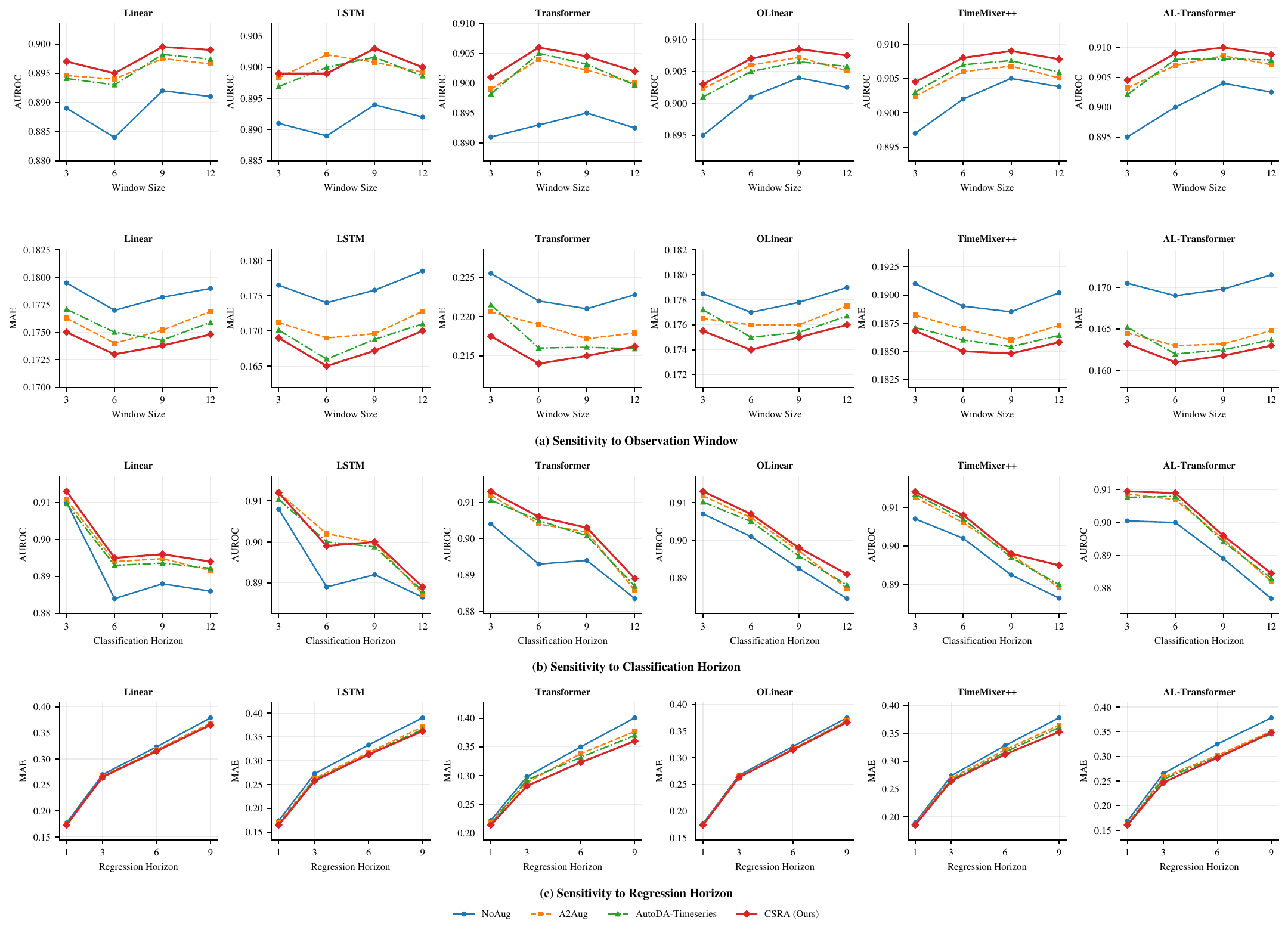}
    \vspace{-3pt}
    \caption{Performance comparison under different observation windows and prediction horizons on six downstream models, where (a), (b), and (c) correspond to varying \(W\), \(H_{\mathrm{cls}}\), and \(H_{\mathrm{reg}}\), respectively. The results show that CSRA exhibits smaller performance degradation under shorter windows and longer horizons.}
    \label{fig:time_sensitivity}
\end{figure*}

To evaluate the robustness of CSRA under different temporal settings, we vary the observation window size \(W\), the classification horizon \(H_{\mathrm{cls}}\), and the regression horizon \(H_{\mathrm{reg}}\), and compare CSRA with NoAug and the two strongest augmentation baselines, A2Aug and AutoDA-Timeseries, across six downstream models. As shown in Fig.~\ref{fig:time_sensitivity}, all methods degrade to different extents as the observation window becomes shorter or the prediction horizon becomes longer. Across most settings, however, CSRA remains the best or among the best methods, with a smaller overall drop in performance.

Figure~\ref{fig:time_sensitivity}(a) shows that reducing the observation window \(W\) consistently weakens performance on both classification and regression, as less historical evidence is available to support prediction. Under these shorter-window settings, CSRA remains competitive across most downstream models, and its advantage becomes more apparent as the window shrinks. This result is aligned with the motivation of CSRA, namely to provide stronger training support when temporal evidence is limited.

A similar trend appears when the classification horizon \(H_{\mathrm{cls}}\) is increased in Fig.~\ref{fig:time_sensitivity}(b). As the lead time becomes longer, classification performance gradually declines, reflecting greater uncertainty and weaker supervision for future risk prediction. Even under these more difficult settings, CSRA remains at or near the top across most downstream models, indicating a clear advantage for longer-horizon risk prediction.

For regression, Fig.~\ref{fig:time_sensitivity}(c) shows that MAE increases more markedly as the regression horizon \(H_{\mathrm{reg}}\) grows. This suggests that continuous state forecasting is more sensitive to prediction distance. Still, CSRA achieves the lowest or near-lowest MAE in most settings, and its relative advantage remains more stable as \(H_{\mathrm{reg}}\) increases. This is particularly relevant for sepsis monitoring, where forecasting future physiological trends becomes progressively harder as the lead time extends.

Overall, these results show that CSRA degrades more gracefully under increasingly stringent temporal settings and maintains more favorable performance under shorter windows and longer horizons.

\subsection{Model Analysis}

To better understand how CSRA perturbs clinical trajectories and how the controller distributes modulation across clinical systems and frequency bands, we provide both controller-level and case-level visualizations.

\begin{wrapfigure}{r}{0.55\columnwidth}
    \vspace{-0.4\baselineskip}
    \centering
    \includegraphics[width=0.53\columnwidth]{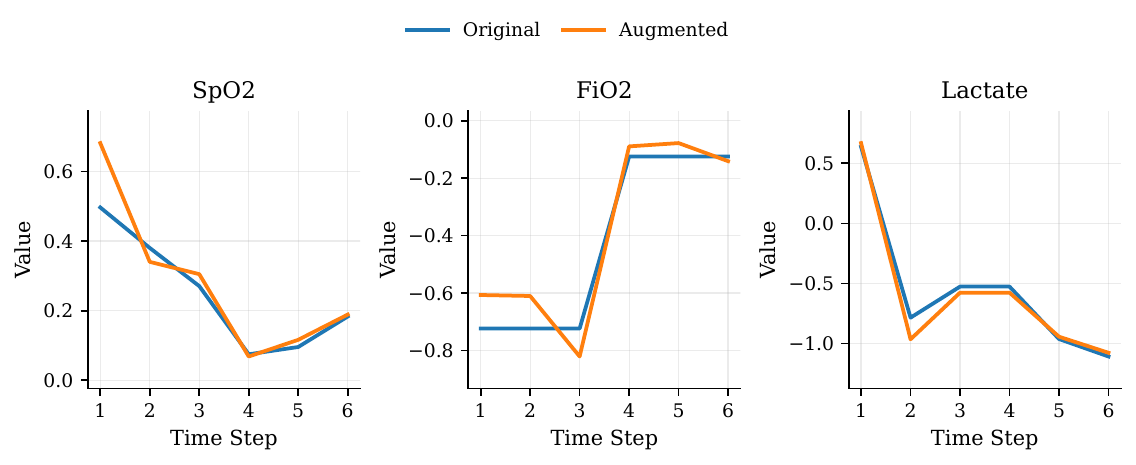}
    \caption{Visualization of the original and augmented trajectories of representative variables from the respiratory (SpO2 and FiO2) and metabolic (Lactate) systems. CSRA preserves the overall temporal trends while introducing structured, system-aware perturbations.}
    \label{fig:case}
    \vspace{-0.7\baselineskip}
\end{wrapfigure}
\begin{figure*}[!th]
    \centering
    \includegraphics[width=0.80\textwidth]{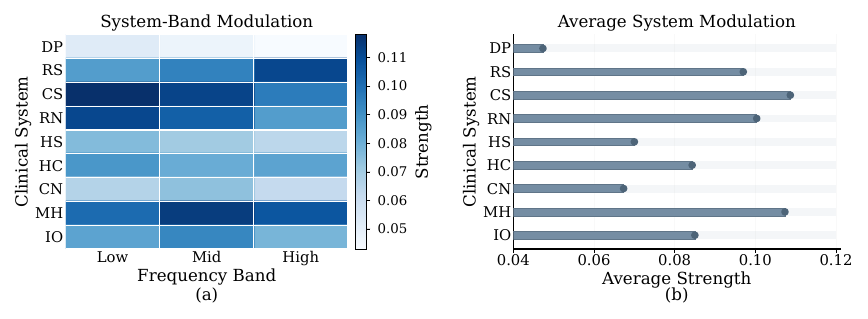}
    \vspace{-3pt}
    \caption{Illustrative visualization of the controller-induced modulation patterns in CSRA. The left panel shows the system--band modulation strength across low-, mid-, and high-frequency bands, and the right panel shows the average modulation strength of each clinical system. DP: Demographic/Profile; RS: Respiratory; CS: Circulatory; RN: Renal; HS: Hepatic; HC: Hematologic/Coagulation; CN: Central Nervous; MH: Metabolic/Homeostasis; IO: Intervention. Different clinical systems exhibit distinct modulation intensities and frequency-band preferences, reflecting the system-aware and structured behavior of the proposed controller.}
    \label{fig:controller_visualization}
\end{figure*}

\textbf{System--Band Modulation Analysis.~} Figure~\ref{fig:controller_visualization} presents the average modulation strength of each clinical system together with the corresponding system--band modulation patterns. Clear differences are observed across systems. In particular, the circulatory, renal, and metabolic/homeostasis systems receive stronger modulation overall, whereas the demographic/profile and central nervous systems remain relatively weak. This pattern is clinically plausible, as sepsis progression is more directly manifested through hemodynamic instability, renal dysfunction, and metabolic disturbance, while demographic variables mainly provide background context. Distinct preferences across frequency bands are also evident. The respiratory system shows relatively stronger high-frequency modulation, whereas the circulatory and renal systems are more pronounced in the low- and mid-frequency bands. These findings indicate that CSRA can adaptively capture both slow-varying trends and short-term fluctuations in a system-dependent manner. Taken together, the results suggest that the controller learns structured and clinically meaningful modulation patterns, rather than applying uniform perturbations across all variables.

\textbf{System-aware Trajectory Visualization.~} Figure~\ref{fig:case} shows the original and augmented trajectories of three representative clinical variables, including two respiratory variables and one metabolic variable. Although the perturbations remain moderate in magnitude, their patterns differ clearly across variables. The augmented trajectories largely preserve the global temporal trend of the original signals, while introducing localized changes in magnitude and short-term dynamics. This result suggests that CSRA performs structured, system-aware augmentation rather than indiscriminately perturbing the input sequence.

\subsection{Clinical Analysis}

To assess the clinical plausibility and potential practical value of the model outputs, we further conducted a clinician-centered subjective evaluation. Given the differences in output form and clinical interpretation between classification and regression, task-specific evaluation criteria were designed for the two tasks.

For classification, clinicians evaluated three aspects: 1) \textbf{state matching}, namely whether the prediction was consistent with the patient’s overall clinical condition within the observation window; 2) \textbf{evolution support}, namely whether the prediction was supported by recent changes in key clinical variables; and 3) \textbf{clinical reference value}, namely whether the output could provide useful auxiliary information for clinical judgment. For regression, clinicians also evaluated three aspects: 1) \textbf{numerical plausibility}, namely whether the predicted values remained within a clinically acceptable range without obviously abnormal or abrupt deviations; 2) \textbf{trend plausibility}, namely whether the predicted future changes followed a clinically reasonable temporal pattern; and 3) \textbf{cross-variable coherence}, namely whether multiple predicted variables jointly formed a consistent and clinically interpretable future state.

In practice, we randomly selected 50 cases from the test set for manual review. For each case, six clinicians were provided with the observed trajectories of key variables within the observation window, the corresponding model outputs, and the relevant prediction targets. To reduce prior bias, all results were presented in anonymized form, and the clinicians were not informed of the source method during scoring. All items were rated on a five-point Likert scale, with higher scores indicating greater perceived clinical value.

\begin{wrapfigure}{r}{0.50\columnwidth}
    \vspace{-0.4\baselineskip}
    \centering
    \includegraphics[width=0.48\columnwidth]{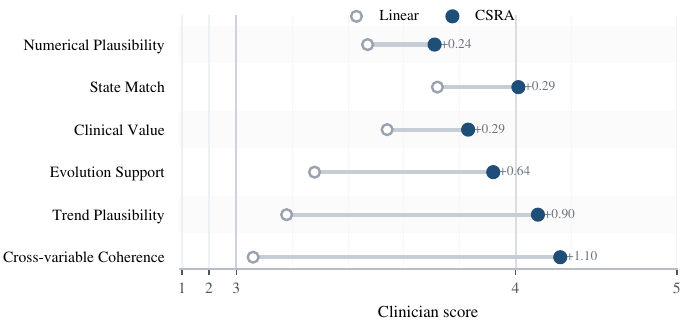}
    \caption{Clinician-centered subjective evaluation of Linear and CSRA on six task-specific dimensions.}
    \label{fig:clinician}
    \vspace{-0.7\baselineskip}
\end{wrapfigure}
As shown in Fig.~\ref{fig:clinician}, CSRA achieved consistently higher subjective scores across all six evaluation dimensions, indicating that its outputs were better aligned with clinicians’ reasoning about patient state and disease progression. The advantage was particularly clear on the regression-related dimensions of trend plausibility and cross-variable coherence, suggesting that CSRA produces predictions that are not only more natural at the level of individual variables, but also more consistent with clinically realistic future trajectories when considered jointly.

\subsection{Generalization on an External Clinical Dataset}

To further evaluate the applicability of the proposed method on other medical datasets, we conducted supplementary experiments on an independent real-world clinical database, \textbf{ZiGongICUinfection}~\cite{ZiGongICUinfection}. This dataset was derived from the infection-related critical care database of Zigong Fourth People’s Hospital in China and consists of multiple heterogeneous clinical tables, including baseline information, laboratory results, medication orders, ICD diagnoses, nursing records, transfer information, and outcomes, covering ICU patients with infection from 2019 to 2020.
\begin{wrapfigure}{r}{0.57\columnwidth}
    \vspace{-0.4\baselineskip}
    \centering
    \includegraphics[width=0.55\columnwidth]{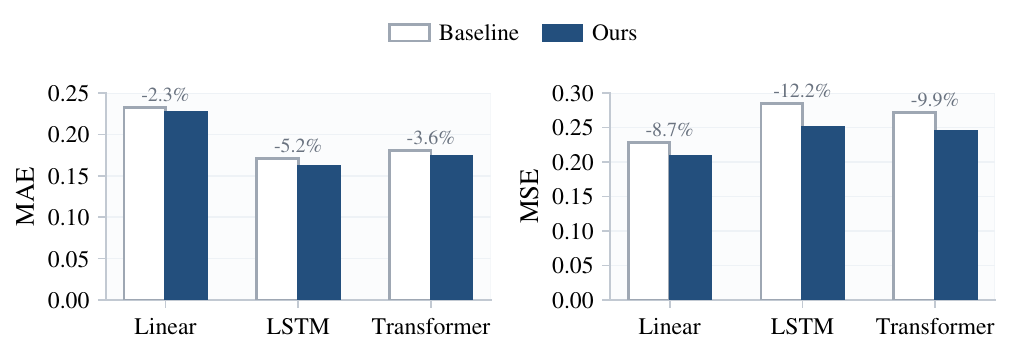}
    \caption{Comparison of regression performance across three downstream  models on the \textbf{ZiGongICUinfection}~\cite{ZiGongICUinfection}, showing that the proposed method remains effective on an external clinical dataset with heterogeneous real-world records.}
    \label{fig:zigong_result}
    \vspace{-0.7\baselineskip}
\end{wrapfigure}

To adapt this dataset to our temporal prediction setting, we first linked multi-source records using the hospital admission identifier and then mapped events from different sources onto a unified timeline to construct patient-level multivariate time series at a fixed 1-hour resolution. During feature curation, we retained structured variables directly related to patient state evolution and clinical interventions, including selected nursing observations and treatment-related events, while excluding text-dependent fields, semantically unstable variables, and fields mainly reflecting device settings or alarm thresholds. For missing values, we first removed variables with missing rates above 80\%, and then applied forward and backward filling within each patient trajectory, with global statistics used as a fallback when necessary. After preprocessing, {ZiGongICUinfection was converted into an hourly aligned patient-level multivariate time-series dataset, containing 2,502 hospital trajectories and 1,761,822 time-step records, with 14 modeling variables at each time step. 

On ZiGongICUinfection, we used an observation window of 24 hours and predicted the variables at the next time step. As shown in Fig.~\ref{fig:zigong_result}, our method achieved consistent performance gains across all three downstream  models, namely Linear, LSTM, and Transformer. Compared with the MIMIC cohort used in the main experiments, ZiGongICUinfection contains finer-grained nursing monitoring information and stronger real-world clinical noise. Even under these more challenging conditions, the method still yielded stable improvements, suggesting that it generalizes well to heterogeneous clinical time series.


%% file: sec/5_lim.tex
\section{Limitations and Future Work}

Despite the encouraging results, several limitations of the current study should be acknowledged. The current framework is developed for regularly sampled time series and does not yet account for irregular clinical observations. In addition, the evaluation is mainly conducted on the MIMIC dataset, and further validation across different patient populations and data collection settings is needed. Moreover, CSRA currently relies on predefined clinical system partitioning and frequency-domain residual augmentation, and its applicability to more complex variable dependencies warrants further study.

Future work will extend CSRA to irregularly sampled clinical time series and explore more adaptive system-aware augmentation strategies. It will also further validate the framework across broader patient populations and clinical settings.

%% file: sec/6_conclusion.tex
\section{Conclusion}

In this paper, we investigated short-window sepsis prediction in challenging temporal settings, where reliable prediction becomes more difficult as the observation window shortens and the prediction horizon extends. To address this issue, we proposed CSRA, a controlled spectral residual augmentation framework for short-window multi-system ICU time series, designed to generate structured and clinically plausible trajectory variations through system-aware modeling and input-adaptive perturbation in the frequency domain. To further stabilize the augmentation process, CSRA is jointly optimized with the downstream predictor using anchor consistency and controller regularization, which help prevent unrealistic augmentation patterns and improve controllability during training.

Experimental results across multiple downstream models show that CSRA consistently outperforms representative baselines on both regression and classification tasks. Moreover, CSRA exhibits smaller performance degradation under shorter observation windows, longer prediction horizons, and smaller training data scales.